\documentclass[letterpaper, 10 pt, conference]{ieeeconf}

\IEEEoverridecommandlockouts                              % This command is only needed if 
\usepackage{epstopdf}
\usepackage{cite}
\usepackage{amsmath,amssymb,amsfonts}
\usepackage{algorithmic}
\usepackage{graphicx}
\usepackage{textcomp}
\usepackage{subcaption}
\usepackage[ruled,vlined]{algorithm2e}
\usepackage{float}
\usepackage{wrapfig}
\usepackage{xcolor}
\usepackage[inkscapelatex=false]{svg}
\usepackage{breqn}
\usepackage{url}

\usepackage{hyperref}
\usepackage{cleveref}

\usepackage{mathtools}
\usepackage{svg}
\usepackage[inkscapelatex=false]{svg}

\usepackage{graphicx}
\usepackage{pgfkeys, pgfsys, pgfcalendar}

\usepackage{pgffor}

\usepackage{multirow}
\usepackage{makecell}

\makeatletter
\AtBeginDocument{%
  \@ifundefined{pgf@sys@addpdfresource@extgs@plain}%
    {\let\pgfutil@addpdfresource@extgs\SomethingProbablyUndefined}{}}
\makeatother
\usepackage{transparent}

\newtheorem{dfn}{Definition}

\newtheorem{thm}{Theorem}

\newtheorem{rem}{Remark}

\makeatletter
\newcommand{\subalign}[1]{%
	\vcenter{%
		\Let@ \restore@math@cr \default@tag
		\baselineskip\fontdimen10 \scriptfont\tw@
		\advance\baselineskip\fontdimen12 \scriptfont\tw@
		\lineskip\thr@@\fontdimen8 \scriptfont\thr@@
		\lineskiplimit\lineskip
		\ialign{\hfil$\m@th\scriptstyle##$&$\m@th\scriptstyle{}##$\crcr
			#1\crcr
		}%
	}
}

\newenvironment{prf}{\textbf{Proof.}}{\hfill$\blacksquare$}

\makeatletter
\newcommand{\pushright}[1]{\ifmeasuring@#1\else\omit\hfill$\displaystyle#1$\fi\ignorespaces}
\newcommand{\pushleft}[1]{\ifmeasuring@#1\else\omit$\displaystyle#1$\hfill\fi\ignorespaces}
\makeatother

\newcommand{\alglinelabel}{%
  \addtocounter{ALC@line}{-1}% Reduce line counter by 1
  \refstepcounter{ALC@line}% Increment line counter with reference capability
  \label% Regular \label
}

\newcommand{\Treplay}{{T_\replay}}										% Replay buffer size
\newcommand{\lrcrit}{\alpha_\crit}										% Critic learning rate
\newcommand{\goaldist}[1][{}]{d_{\G#1}}									% Distance to goal set

\definecolor{gray}{rgb}{0.5, 0.5, 0.5}

\DeclareMathOperator*{\argmin}{\text{arg\,min}}							% Argmin
\newcommand{\nrm}[1]{\left\lVert#1\right\rVert}							% Norm
\newcommand{\E}[2][{}]{\mathbb E_{#1}\left[#2\right]}					% Mean
\newcommand{\PP}[1]{\mathbb P\left[#1\right]}							% Probability
\newcommand{\low}{{\text{low}}}											% Lower bound
\newcommand{\up}{{\text{up}}}											% Upper bound
\newcommand{\ra}{\rightarrow}											% Right arrow
\newcommand{\la}{\leftarrow}											% Left arrow
\newcommand{\ie}{\unskip, i.\,e.,\xspace}								% That is
\newcommand{\eg}{\unskip, e.\,g.,\xspace}								% For example
\newcommand{\sut}{\text{s.\,t.\,}}										% Such that or subject to
\newcommand{\Z}{{\mathbb{Z}}}											% Set of integer numbers
\newcommand{\R}{{\mathbb{R}}}											% Set of real numbers
\newcommand{\blue}[1]{\textcolor{blue}{#1}}
\definecolor{dgreen}{rgb}{0.0, 0.5, 0.0}

\newcommand{\state}{s}													% State (as vector)
\newcommand{\State}{S}													% State (as random variable)
\newcommand{\states}{\mathbb S}											% State space
\newcommand{\action}{a}													% Action (as vector)	
\newcommand{\Action}{A}													% Action (as random variable)
\newcommand{\actions}{\mathbb A}										% Action space
\newcommand{\policy}{\pi}												% Policy (as function or distribution)
\newcommand{\policies}{\Pi}												% Policy space
\newcommand{\transit}{p}												% State transition map
\newcommand{\cost}{c}													% Cost (as vector)
\newcommand{\Value}{V}													% Value
\newcommand{\W}{\ensuremath{\mathbb{W}}}								% Weight space
\newcommand{\crit}{{\text{crit}}}										% Critic abbreviation
\newcommand{\G}{\ensuremath{\mathbb{G}}}								% Attractor (goal set)
\newcommand{\K}{\ensuremath{\mathcal{K}}\xspace}						% Class kappa
\newcommand{\KL}{\ensuremath{\mathcal{KL}}\xspace}						% Class kappa-ell
\newcommand{\Kinf}{\ensuremath{\mathcal{K}_{\infty}}\xspace}			% Class kappa-infinity
\newcommand{\KLinf}{\ensuremath{\mathcal{KL}_{\infty}}\xspace}			% Class kappa-ell-infinity
\newcommand{\loss}{\mathcal L}											% Loss
\newcommand{\replay}{\mathcal R}										% Experience replay
\newcommand{\spc}{{\,\,}}												% White space to be used in logical formulas

\begin{document}

\title{A novel agent with formal goal-reaching guarantees: an experimental study with a mobile robot}
%\author{Anonymous}
\author{Grigory Yaremenko$^{1}$, Dmitrii Dobriborsci$^{2}$, Roman Zashchitin$^{2}$, \\ Ruben Contreras Maestre$^{2}$, Ngoc Quoc Huy Hoang$^{2}$, Pavel Osinenko$^{1}$
\thanks{Grigory Yaremenko and Dmitrii Dobriborsci contributed equally.}
\thanks{$^{1}$Skolkovo Institute of Science and Technology}
\thanks{$^{2}$Deggendorf Institute of Technology, Technology Campus Cham}
\thanks{Corresponding author: P. Osinenko, email: \texttt{p.osinenko@gmail.com}.}
%\thanks{Pavel Osinenko is the corresponding author.}
}

\maketitle

%%%%%%%%%%%%%%%%%%%%%%%%%%%%%%%%%%%%%%%%%%%%%%%%%%%%%%%%%%%%%%%%%%%%%%%%%%%%%%%%%%%%%%%%%%%%
%%%%%%%%%%%%%%%%%%%%%%%%%%%%%%%%%%%%%%%%%%%%%%%%%%%%%%%%%%%%%%%%%%%%%%%%%%%%%%%%%%%%%%%%%%%%

% \green{CHANGES RELATED TO FIXES OF OVERSHOOT MARKED GREEN}

%\twocolumngrid

\begin{abstract}
Reinforcement Learning (RL) has been shown to be effective and convenient for a number of tasks in robotics. 
However, it requires the exploration of a sufficiently large number of state-action pairs, many of which may be unsafe or unimportant. 
For instance, online model-free learning can be hazardous and inefficient in the absence of guarantees that a certain set of desired states will be reached during an episode. 
An increasingly common approach to address safety involves the addition of a shielding system that constrains the RL actions to a safe set of actions. 
In turn, a difficulty for such frameworks is how to effectively couple RL with the shielding system to make sure the exploration is not excessively restricted. 
%In this paper, we frame safety as a differentiable robust-control-barrier-function layer in a modelbased RL framework. Moreover, we also propose an approach to
%modularly learn the underlying reward-driven task, independent
%of safety constraints. We demonstrate that this approach both
%ensures safety and effectively guides exploration during training
%in a range of experiments, including zero-shot transfer when the
%reward is learned in a modular way.
This work presents a novel safe model-free RL agent called Critic As Lyapunov Function (CALF) and showcases how CALF can be used to improve upon control baselines in robotics in an efficient and convenient fashion while ensuring guarantees of stable goal reaching. The latter is a crucial part of safety, as seen generally.
With CALF all state-action pairs remain explorable and yet reaching of desired goal states is formally guaranteed.
Formal analysis is provided that shows the goal stabilization-ensuring properties of CALF and a set of real-world and numerical experiments with a non-holonomic wheeled mobile robot (WMR) TurtleBot3 Burger confirmed the superiority of CALF over such a well-established RL agent as proximal policy optimization (PPO), and a modified version of SARSA in a few-episode setting in terms of attained total cost.
%Online means that in each learning episode, the said environment is stabilized.
%This, as demonstrated in a case study with a mobile robot simulator, greatly improves the overall learning performance.
%The base actor-critic scheme of CALF is analogous to SARSA.
%The latter did not show any success in reaching the target in our studies.
%However, a modified version thereof, called SARSA-m here, did succeed in some learning scenarios.
%Still, CALF greatly outperformed the said approach.
%CALF was also demonstrated to improve a nominal stabilizer provided to it.
%In summary, the presented agent may be considered a viable approach to fusing classical control with reinforcement learning.
%Its concurrent approaches are mostly either offline or model-based, like, for instance, those that fuse model-predictive control into the agent.
\end{abstract}

%%%%%%%%%%%%%%%%%%%%%%%%%%%%%%%%%%%%%%%%%%%%%%%%%%%%%%%%%%%%%%%%%%%%%%%%%%%%%%%%%%%%%%%%%%%%
%%%%%%%%%%%%%%%%%%%%%%%%%%%%%%%%%%%%%%%%%%%%%%%%%%%%%%%%%%%%%%%%%%%%%%%%%%%%%%%%%%%%%%%%%%%%
\section{Introduction}
\label{sec_sota}

In recent years, model-free Reinforcement Learning (RL) techniques have gained popularity in mobile robotics, especially in environments where optimality and efficiency are critical.
Great effort was put into deriving formal guarantees in RL, which still remains one of the greatest challenges.
A large number of works within the subject of RL with guarantees has been conducted in the context of reaching a safe set (a goal) or avoiding low-reward areas (\cite{buhrer2023multiplicativevaluefunctionsafe, Yang2021WCSACWS, wen2020safereinforcementlearningautonomous}).

%For instance, there are works that describe ways of efficiently minimizing constraint violations in a similar fashion to minimizing costs \cite{buhrer2023multiplicativevaluefunctionsafe, Yang2021WCSACWS, wen2020safereinforcementlearningautonomous}.
%This however is not suitable for cases when online data-driven learning is needed or when formal guarantees of constraint satisfaction are required.
%One example of this is \cite{buhrer2023multiplicativevaluefunctionsafe}, which introduces a safe model-free RL algorithm to enhance safety and reward efficiency in robot navigation tasks. By integrating a multiplicative value function that incorporates safety and reward critics, the RL agent can predict the likelihood of constraint violations and adjust its strategy accordingly. 
%This method was successfully tested on real-world robot navigation tasks using raw sensor data, such as LiDAR scans, and demonstrated efficient zero-shot sim-to-real transfer. 
%\cite{Yang2021WCSACWS} proposed the Worst-Case Soft Actor Critic (WCSAC) algorithm, which integrates a safety critic that estimates worst-case scenarios using conditional value-at-risk (CVaR). WCSAC is designed to ensure that the robot’s policies remain safe even in high-risk situations, balancing risk aversion with task performance. The method has been validated in dynamic environments such as the Safety Gym benchmark, demonstrating its effectiveness in optimizing safe policies under risk constraints.
For instance, \cite{emam2022safereinforcementlearningusing} combined Robust Control Barrier Functions (RCBFs) with RL to ensure exploration in high-reward areas and control in continuous tasks.
By integrating a differentiable RCBF safety layer into the Soft Actor-Critic (SAC) framework, this method provides real-time guarantees while optimizing the robot's navigation behavior.
The main innovation here is the ability to maintain guarantees during exploration without sacrificing learning performance.
This approach ensures that the learned policies balance safety and task performance effectively, however for a general system the design of a control barrier function can be a highly creative and sophisticated procedure.

A method of Projection-Based Constrained Policy Optimization (PCPO) that optimizes reward functions while ensuring that certain state constraints are met through a projection step was presented in \cite{yang2020projectionbasedconstrainedpolicyoptimization}.
By iteratively improving the policy and projecting it back onto the constraint set, PCPO achieves more efficient constraint satisfaction compared to therein studied baselines.
%The algorithm demonstrated superior performance across several control tasks, showing three times fewer constraint violations while maintaining high reward efficiency. This is particularly useful in robotics applications like autonomous navigation, where balancing exploration and safety is critical.
%Similarly, a Parallel Constrained Policy Optimization (PCPO) was introduced by \cite{wen2020safereinforcementlearningautonomous}. This advanced model-free RL algorithm ensures safety during autonomous driving tasks. The algorithm builds upon actor-critic architectures by incorporating a risk function that estimates potential hazards and enforces safety constraints throughout the learning process. This is further enhanced by the use of synchronized parallel learners, which explore different state subspaces, speeding up learning and improving policy updates.
The work \cite{chow2019lyapunovbasedsafepolicyoptimization} also contributed by presenting a method for optimizing policies under the framework of constrained Markov decision processes (CMDPs) using Lyapunov functions for formal guarantees of goal stabilization at every stage of training.
By taking advantage of deep deterministic policy gradient (DDPG) and proximal policy optimization (PPO), the approach balances the robot's need to explore while adhering to constraints during navigation.
This technique has been successfully applied in simulations and real-world indoor robot navigation tasks, offering a robust solution to safe and efficient exploration in dynamic environments, however, naturally, it is implied that a Lyapunov function is priorly known.

In line with this, \cite{han2020actorcriticreinforcementlearningcontrol} proposed an actor-critic RL framework that guarantees system stability, a key concern for real-world robotic applications, especially in nonlinear, high-dimensional control tasks. 
Their method uses Lyapunov's method to ensure the stability of learned policies, enabling the systems to recover when uncertainties interfere.
%This approach has been shown to enhance robustness and control performance in various simulated robot tasks. 

Similarly, \cite{huh2020safereinforcementlearningprobabilistic} introduced a model-free RL method that combines probabilistic reachability analysis and Lyapunov-based techniques to ensure safety. 
This method constructs a Lyapunov function during policy evaluation to provide guarantees and guide exploration, gradually expanding the set of states the robot can explore without exceeding stated constraints, however this only provides probabilistic constraints.
%The approach is particularly effective in high-dimensional systems, leveraging deep RL with actor-critic algorithms to efficiently balance exploration and safety in continuous control tasks.
Further advancements include \cite{xiong2022modelfreeneurallyapunovcontrol}, which presented a model-free neural Lyapunov control approach that combines the flexibility of deep RL with the rigorous guarantees of Lyapunov functions.
This proposed method, which co-learns a Twin Neural Lyapunov Function (TNLF) alongside a control policy, provides runtime safety monitoring for robot navigation by guiding robots through collision-free trajectories using raw sensor data such as LiDAR, though it must be pointed out that the approach is not completely risk-free and it is only suitable for a particular set of navigation tasks.

\textbf{Two common patterns} can be traced: 
\begin{enumerate}
\item No guarantees provided, but no prior knowledge required,
\item Rigorous guarantees provided, but prior knowledge or ad hoc derivations are required.
\end{enumerate}
This is explained by the fact that having the full advantages of both likely poses an intractable problem.
Indeed, quadratically constrained quadratic programming is NP-hard \cite{Pardalos1991} and hence so are nonlinear optimal control problems, feasibility problems, hard-constrained MDPs and other formalizations that are characteristic of RL with guarantees in robotics, which means that likely all algorithms that can produce provably safe policies in an end-to-end fashion are either computationally inefficient or only suitable for a particular subset of problems.

This work contributes to state of the art in approaches of type 2.
At present type 2 can be further subdivided into the following categories: 
\begin{itemize}
\item Shield-based RL, which uses a pre-designed action filter to avoid unsafe actions.
Approaches like this normally involve either ad hoc design, the knowledge of a Lyapunov function or sophisticated formal verification.
Exemplary works are \eg \cite{Tan2020DeductiveStabi,Platzer2008Keymaerahybrid,Fulton2018Safereinforcem} that utilize specific facts about the model.
\item RL fused with model-predictive control \eg \cite{Lowrey2018Planonlinelea,East2020InfiniteHorizo,Reddy2019LearningHuman,Finn2017Deepvisualfor,Karnchanachari2020PracticalReinf,Asis2020FixedHorizonT}.
Such approaches draw on the MPC's well-established ability to ensure closed-loop stability via techniques like terminal costs and constraints.
Proposals such as the one by Zanon et al. \cite{Zanon2020SafeReinforcem,Zanon2019Practicalreinf} embed robust MPC within reinforcement learning to maintain safety.
%\cite{Zanon2020SafeReinforcem,Zanon2019Practicalreinf,Koller2018LearningBased,Berkenkamp2017SafeModelbase,Berkenkamp2019Safeexploratio,Oh2023QMPCstable}.

\item Lyapunov-based reinforcement learning such as \cite{Chow2018Lyapunovbased,Berkenkamp2017SafeModelbase,Jeddi2023Memoryaugmente,Han2020Actorcriticre,Chang2021Stabilizingneu}, which  typically makes use of a known Lyapunov or control barrier functions to ensure that the target is reached and/or high-penalty states are avoided.
Other approaches of this kind may instead attempt to construct a Lyapunov function during offline learning that certifies the obtained policy \cite{Berkenkamp2017SafeModelbase}.
\end{itemize}

\emph{Each of the above involves either prior knowledge of a Lyapunov function, ad hoc formal derivation\footnote{For instance, deriving a solution to the HJB equation, solving the co-state ODE for Pontryagin's maximum principle, etc.}, post hoc verification or extra assumptions}.

The present paper proposes an RL approach with guarantees, having neither of the above stated limitations.

\subsection{Contribution}

An approach to RL with formal guarantees is proposed.
The proposed approach is claimed to efficiently improve a given baseline policy while preserving the guarantees of reaching a set of goal states.
The said claims are confirmed with a formal result (see \Cref{thm_calfstabmean}) and an experimental evaluation with a non-holonomic wheeled mobile robot (WMR) TurtleBot3 Burger (see \cref{sec_results}). 
The proposed approach fills the research gap of finding a more practical and convenient alternative to RL with guarantees that requires the knowledge of a Lyapunov function.
The present paper extends the results in \cite{Osinenko2024} to the stochastic setting, significantly generalizing original formal results and adding considerable flexibility to the implementation.
Importantly, the real-world experiments in mobile robotics for the first time verify the applicability of an approach of this kind in an applied setting. 

%\red{Remark: The problem of goal-directed navigation for mobile robots can indeed be addressed using conventional planning algorithms such as Rapidly-exploring Random Trees (RRT), Dijkstra’s algorithm, and A*. 
%However, this study introduces a novel reinforcement learning (RL) agent specifically designed for this task. 
%Our primary focus is on the development and evaluation of the RL agent’s capabilities. 
%Therefore, we do not benchmark our RL approach against traditional planning algorithms, as the scope of this research is centered on advancements in reinforcement learning rather than a comparative analysis with established planning methods.}
%
%\red{OR While conventional path planning algorithms such as Rapidly-exploring Random Trees (RRT), Dijkstra’s algorithm, A*, and potential fields are well-established for navigating robots to target positions, the focus of this paper is not on path planning per se. 
%Instead, our research emphasizes the novel application of reinforcement learning (RL) to enable a robot to learn and adapt its behavior to reach a goal position autonomously. 
%The primary objective is to explore the efficacy of RL in uncertain environments, where traditional path planning methods may not be as effective or adaptable. 
%Therefore, a direct comparison with conventional planning algorithms is beyond the scope of this study. 
%Our contribution lies in demonstrating the potential of RL to enhance robotic autonomy and decision-making in complex scenarios.}

\section{Background and problem statement}
\label{sec_problem}

\textbf{Notation:}
%We will use Python-like array notation \eg $[0:T] = \{0, \dots, T-1\}$ or $\state_{0:T} = \{\state_0, \dots, \state_{T-1}\}$.
Spaces of class kappa, kappa-infinity, kappa-ell and kappa-ell-infinity functions are denoted $\K, \Kinf, \KL, \KLinf$, respectively.
These are scalar monotonically increasing functions, zero at zero, and, additionally, tending to infinity in case of $\Kinf$. 
%The subscript $\ge 0$ in number set notation will indicate that only non-negative numbers are meant.
%The notation ``$\cl{\bullet}$'', when referring to a set, will mean the closure.
%We treat $\deltau$ as a single symbol denoting the action step size \ie the physical time elapsed between each two consecutive $\action_t$ and $\action_{t+1}$.
%We denote modulo division of the first argument by the second as ``$\moddiv{\bullet}{\bullet}$''.

\subsection{Problem statement}
%\section{Background and problem statement}
\label{sec_problem}

Consider the following Markov decision process (MDP):
\begin{equation}
	\label{eqn_mdp}
	\left(\states, \actions, \transit, \cost \right),
\end{equation}
where:
\begin{enumerate}
\item $\states$ is the \textit{state space}, assumed as a finite-dimensional Banach space of all states of the given environment;
\item $\actions$ is the \textit{action space}, that is a set of all actions available to the agent, assumed to be a compact topological space;
\item $\transit : \states \times \actions \times \states \ \rightarrow \ \R$ is the \textit{transition probability density function} of the environment, that is such a function that $\transit(\bullet \mid \state_{t}, \action_{t})$ is the probability density of the state $s_{t + 1}$ at step $t+1$ conditioned on the current state $\state_{t}$ and current action $\action_{t}$;
\item $\cost : \states \times \actions \rightarrow \mathbb{R}$ is the cost function of the problem, that is a function that takes a state $\state_{t}$ and an action $\action_{t}$ and returns the immediate cost $\cost_{t}$ incurred upon the agent if it were to perform action $\action_{t}$ while in state $\state_{t}$.
\end{enumerate}

Let $(\Omega, \Sigma, \mathbb P)$ be a probability space underlying \eqref{eqn_mdp}, and $\mathbb E$ be the respective expected value operator.
The problem of reinforcement learning is to find a policy $\policy$ from some space of admissible policies $\policies$ that minimizes 
\begin{equation}
	\label{eqn_value}
	\Value^{\policy}(\state) := \E[\Action \sim \policy]{\sum_{t = 0}^{\infty}\gamma^{t}\cost(\State_{t}, \Action_{t}) \mid \State_0=\state}, \state \in \states	
\end{equation}
for some $\gamma \in (0, 1]$ called a discount factor.
The problem may be concerned with a designated initial state $\state$, a distribution thereof, or even the whole state space.
% A policy may be taken as a probability density function or as an ordinary function (cf. Markov policy).
The policy $\policy^*$ that solves the stated problem is commonly referred to as the optimal policy.
An agent will be referred to as a finite routine that generates actions from the observed states.

\begin{dfn}
A policy $\policy_0 \in \policies$ is said to satisfy the $\eta$-improbable \textbf{goal reaching property}, $\eta \in [0,1)$, if
\begin{equation}
	\label{eqn_introstab}
		\forall \state_0 \in \states \spc \PP{\goaldist(\State_t) \xrightarrow{t \ra \infty} 0 \mid \Action_t \sim \policy^0_{t}(\State_{t})} \ge 1 - \eta.
\end{equation}
\end{dfn}
Here $\goaldist(\State_t) \xrightarrow{t \rightarrow \infty} 0$ denotes that $0$ is a limit point of $\goaldist(\State_t)$.
That is, for all starting states, $\policy_0$ ensures that the probability of failing to reach the goal is no greater than $\eta$. 

The question posed by the present paper is the following one:
\begin{center}
\textit{If one knows a goal reaching policy $\pi^{0}$, how does one \underline{efficiently} improve upon it \underline{from data} in a way that guarantees to \underline{preserve the goal-reaching property}?}
\end{center}

There is a wealth of cases in robotics, when a simplistic yet working policy is well known, but determining a truly efficient one is significantly more challenging.
Consider for instance support-polygon-only walking vs. jumping or sprinting locomotion.
In a scenario like this it is desirable to have some means of getting from simple to efficient conveniently, reliably and consistently.
Another example is navigation of a robot into a safe goal with avoidance of high-penalty areas\eg due to hazards.
A naive, yet goal-reaching controller, may be designed with ease, yet RL, that would account for high-penalty areas, may help avoiding them.
At the same time, with no guarantees, such an RL agent may waste numerous episodes before learning to achieve the said safe goal.

\section{Approach}
\label{sec_approach}

In order to solve the aforementioned problem the present paper proposes an approach named Critic As A Lyapunov Function (CALF) a particular implementation of which is described in Algorithm \ref{alg_calfq}.

In essence CALF is merely a system of Lyapunov-like constraints for critic updates coupled with a recovery procedure for when no feasible solution could be found during such an update.

The idea is that if constraint in line \ref{algline_calfcriticupdate}: are satisfied, then the resulting Lyapunov-like relation would
ensure that $\mathbb G$ is reached, much akin to how the existence of a Lyapunov function certifies stability.

On the other hand if the constraints fail, the provided baseline $\policy^0$ is invoked, ceasing control until the environment is driven to a narrower area around $\mathbb{G}$, where a critic update is once again feasible. Indeed, it can observed that the constraints in \ref{algline_calfcriticupdate}: will always be feasible in a sufficiently small neighbourhood of $0$, provided that the critic architecture is sufficiently flexible and that $\hat Q^{w^\dagger}(\state^\dagger, \action^\dagger) > {\bar \nu}$.

Unlike the shielding approaches available in the literature CALF \underline{does not make any state-action pairs unexplorable}. Although a naive decision may result in $\pi_0$ ceasing control immediately after, the outcome of having made the decision can always be explored. The amount of ``freedom'' can also be tuned by modifying $\bar \nu$.
Smaller values of $\bar \nu$ correspond to lighter restrictions on learning and naive exploration.
In fact any finite state-action sequence can be explored without interruptions given that $\bar \nu$ is sufficiently small and the critic architecture is reasonable\eg if the output layer has a bias or if the critic is at least capable of approximating constant functions locally.

It is important to point out that the goal reaching property will be preserved regardless of the values assigned to hyper-parameters (as long as $\policy^0$ is goal-reaching and $\bar \nu > 0$).

\subsection{Lyapunov-like constraints and implementation details}

%Along $w^\dagger$, the state-action pair is also ``stored'', namely, $\state^\dagger, \action^\dagger$.
%The Lyapunov-like constraints are checked relative to $\state^\dagger, \action^\dagger, w^\dagger$.
%This trick enables to safely combine the agent with $\policy_0$ so as to retain the stabilization guarantee of the latter.
%We use the Q-function to make the overall approach model-free.
%The actor loss $\loss^\act$ may be taken equal the current critic with last ``successful'' weights $w^\dagger$, namely, $\loss^\act(\action) = \hat Q^{w^\dagger}(\state_t, \action)$.
%Any augmentation of the loss \eg penalties or entropy terms, may be included into  $\loss^\act$, there is no restriction.
CALF offers a great deal of implementation flexibility.
One can choose a custom critic loss, a custom action update routine, a custom exploration policy and even a custom optimizer so long as constraint satisfaction or the absence thereof is being monitored.

\emph{These choices will not affect CALF's guarantees of goal reaching.}
For the rigorous statement and proof of this result see \cref{sec_theorems}.

%Furthermore, one may choose to take an $\eps$-greedy action.
 
Regarding the $\Kinf$-bounds (see Algorithm \ref{alg_calfq}, line 7), a reasonable choice of $\hat \kappa_\low$, $\hat \kappa_\up$ would be
\begin{equation}
	\label{eqn_calfq_quadkapps}
	\hat \kappa_\low(\bullet) = C_\low \bullet^{2}, \ \hat \kappa_\up(\bullet) = C_\up \bullet^{2}, \spc 0 < C_\low < C_\up.
\end{equation}
% For the decay constraint, one may also take a quadratic rate or simply a constant $\bar \nu > 0$.
Overall, the hyper-parameters $\hat \kappa_\low$, $\hat \kappa_\up$ and $\bar \nu$ determine a trade-off between freedom of learning and worst-case-scenario reaching time of the goal.
Both $\frac{C_\low}{C_\up}$ and $\bar \nu$ determine the trade-off between freedom of exploration and worst-case reaching time.

Now, the critic loss $\loss^\crit$ (see Algorithm \ref{alg_calfq}, line 7) may be chose according to the user's preference, one may for instance take a TD(1) on-policy loss as per:
\begin{equation}
	\label{eqn_TD1}
	\begin{aligned}
		\loss^\crit(w) = & \sum_{k = 0}^\Treplay \big( \hat Q^w(\state_{t_k}, \action_{t_k}) - \cost(\state_{t_k}, \action_{t_k}) - \\
		            & \quad \hat Q^{w^\dagger}( \state_{t_{k+1}}, \action_{t_{k+1}} ) \big)^2 + \lrcrit^{-2}\nrm{w - w^\dagger}^2.
	\end{aligned}
\end{equation}
The regularization term $\lrcrit^{-2}\nrm{w - w^\dagger}^2$ is redundant if gradient descent based minimization is used, since one could simply set a learning rate $\lrcrit$ as opposed to penalizing the displacement of weights.
Notice that the choice of the critic loss (or learning rate) does not affect the formal guarantee of reaching $\mathbb G$, although the quality of the learning may be affected.
Finally, \eqref{eqn_TD1} is akin to the critic loss of SARSA due to its on-policy character, but this is not necessary, an off-policy loss may be used as well \eg with greedy actions instead of $\action_{t_{k+1}}$ in \eqref{eqn_TD1}.

%The stabilizing constraints read:
%\begin{equation}
%	\label{eqn_qstabconstr}
%	\begin{aligned}
%		 & \hat Q^w(\state_t, \action_t) - \hat Q^{w^\dagger}(\state^\dagger, \action^\dagger) \leq -{\bar \nu}, \\
%		 & \hat \kappa_\low (\nrm{\state_t}) \leq \hat Q^{w}(\state_t, \action_t) \le \hat \kappa_\up(\nrm{\state_t}).
%	\end{aligned}
%\end{equation}

%\green{CRITIC MUST INIT AT LEAST WITHIN THE KAPPAS! THIS IS IMPORTANT}

\begin{algorithm}
	\begin{algorithmic}[1]
	\STATE \textbf{Input}: ${\bar \nu} > 0, \hat \kappa_\low, \hat \kappa_\up, \policy^0: \text{baseline}$
	\STATE \textbf{Initialize}: $\state_0, \action_0  \sim \policy^0_0(\state_0), w_0$ \sut
	\[
		\hat \kappa_\low (\nrm{\state_0}) \le \hat Q^{w_0}(\state_0, \action_0) \le \hat \kappa_\up(\nrm{\state_0})
	\]
	\STATE $w^\dagger \gets w_{0}, \state^\dagger \gets \state_0, \action^\dagger \gets \action_{0}$ %, \action_0 \gets \policy_0(\state_0)$
	\FOR {$t := 1, \dots \infty$}
		\STATE Take action $\action_{t -1}$, get state $\state_t$
	   	\STATE Update action: $\action^* \gets \argmin\limits_{\action \in \actions} \hat Q^{w^\dagger}(\state_t, \action)$
%	   	or \eg $\eps$-greedy 		
		\STATE Try critic update \alglinelabel{algline_calfcriticupdate}
%		\vspace*{-\baselineskip}
		\[
		\begin{array}{lll}
	 		w^*  \gets  &  \hspace{-5pt} \argmin\limits_{w \in \W} \loss^\crit(w) \\
				& \hspace{-8pt} \sut \; \hat Q^w(\state_t, \action_t) - \hat Q^{w^\dagger}(\state^\dagger, \action^\dagger) \leq -{\bar \nu}, \\
				& \hspace{-8pt} \phantom{\sut \;} \hat \kappa_\low (\nrm{\state_t}) \le \hat Q^{w}(\state_t, \action_t) \le \hat \kappa_\up(\nrm{\state_t})
		\end{array}
		\]
%		\vspace*{-\baselineskip}
		\IF{ solution $w^*$ found} \alglinelabel{algline_calfcheck} 
		\STATE $\state^\dagger \gets \state_t, \action^\dagger \gets \action^*, w^\dagger \gets w^*$
		\ELSE
		\STATE $\action_t \gets \text{sampled from } \policy^0_{t}(\state_t)$	\alglinelabel{algline_recovery}
		\ENDIF	
	\ENDFOR
	\end{algorithmic}
	\caption{Critic as Lyapunov function (CALF) algorithm, model-free, action-value based}
	\label{alg_calfq}
\end{algorithm}

\begin{rem}
\label{rem_actasmodel}
An actor model $\policy^\theta$ \eg as probability distribution, with weights $\theta$ may be employed instead of direct actions $a_t$ in Algorithm \ref{alg_calfq}.
\end{rem}

\subsection{Transfer of knowledge}

There are several ways in which CALF's knowledge transferring capabilities can be explained. 
First of all, non-goal-reaching policies are excluded from consideration, thus significantly narrowing down the search-space. 
% The critic would in essence only be learning about  goal-reaching policies and thus it can be speculated that more of this learning will remain relevant in later epochs even when the policy changes significantly as compared to earlier epochs.
There is also a more complex consideration: when for some sequence of $N$ states and actions the nominal policy $\pi_{0}$ is being invoked, the critic will be updated to reflect that the state at which the sequence began is associated with the cost produced by the nominal policy\ie
$$
\hat Q(\state^{\dagger}, \action^{\dagger}) \gets \sum_{i = 0}^{N - 1}\cost(\state_{i}, \action_{i}) + \hat Q(\state_{N}, \action_{N}), \ \action_{i} \sim \pi_{t^{\dagger} + i}^{0}.
$$ 
This grants a solid upper bound that can be further improved upon in later episodes, much akin to how the application of the Bellman operator improves upon earlier obtained upper bounds on the cost-to-go function.

Therefore, the following can be argued:
\begin{center}
\textit{Not only does CALF \underline{preserve the goal reaching property}, but it also \underline{enables the transfer of knowledge} from the baseline to the agent being trained.}
\end{center}

The latter claim is consistent with the outcomes of experiments that demonstrate a considerable superiority of CALF over both $\policy^{0}$ and PPO.

%\subsection{Modified SARSA}
%\label{sec_sarsam}
%
%The new CALF agent was benchmarked via its immediate reinforcement learning alternative, \blue{\textit{State–action–reward–state–action} (SARSA)}, which is essentially like Algorithm \ref{alg_calfq} prescribes (with on-policy critic loss), but with the Lyapunov-like constraints and the $\policy_0$ removed.
%In our case studies with a mobile robot, we observed such a plain SARSA failed to drive the robot to the target area within any reasonable number of learning iterations.
%To help SARSA succeed, we slightly modified it, namely, we retained the $w^\dagger$-mechanism \ie we used $w^\dagger$ in the critic loss (see Algorithm \ref{alg_calfq}, line 7).
%We did not \textit{enforce} the Lyapunov-like constraints in the optimization though.
%We only checked, whether those constraints were satisfied post factum and updated $w^\dagger$ accordingly as in CALF.
%Such a modification turned out to help SARSA reach the target in some learning runs.
%This algorithm will further be referred to as SARSA-m.

%%%%%%%%%%%%%%%%%%%%%%%%%%%%%%%%%%%%%%%%%%%%%%%%%%%%%%%%%%%%%%%%%%%%%%%%%%%%%%%%%%%%%%%%%%%%
%%%%%%%%%%%%%%%%%%%%%%%%%%%%%%%%%%%%%%%%%%%%%%%%%%%%%%%%%%%%%%%%%%%%%%%%%%%%%%%%%%%%%%%%%%%%
\section{Analysis}
\label{sec_theorems}

Let $\State^{\policy}_t(\state_0)$ denote the state trajectory emanating from $\state_0$ under a policy $\policy \in \policies$ \ie $\State^{\policy}_0(\state_0) = \state_0$ and for $t > 0$ it holds that $\State^{\policy}_{t}(\state_0) \sim \transit(\bullet \ | \ \State^{\policy}_{t - 1}(\state_0), \Action_{t - 1}(\state_0))$, with $\Action_{t - 1}(\state_0) \sim \policy_{t - 1}(\bullet \mid \State^{\policy}_{t - 1}(\state_0))$.

The main result on preservation of the goal-reaching property is formulated in Theorem \ref{thm_calfstabmean}.

%\green{Statement polished}
%
%\green{Remark: it is only reasonable to bound Q via two kappas, that depend purely on state, only if the action space is compact for each state?
%}
\begin{thm}
	\label{thm_calfstabmean}
	Consider the problem \eqref{eqn_value} under the MDP \eqref{eqn_mdp}.
	Let $\policy^0 \in \policies_0$ have the following goal reaching property for $\G \subset \states$ \ie
	\begin{equation}
	\label{eqn_stabilization}
		\forall \state_0 \in \states \spc \PP{\goaldist(\State^{\policy^0}_t(\state_0)) \xrightarrow{t \ra \infty} 0} \ge 1 - \eta, \eta \in [0,1).
	\end{equation}
	Let $\policy_t$ be produced by Algorithm 1 for all $t \ge 0$.
	Then, a similar goal reaching property is preserved under $\policy_t$ \ie
	\begin{equation}
	\label{eqn_stabilizationcalf}
		\forall \state_0 \in \states \spc \PP{\goaldist(\State^{\policy}_t(\state_0)) \xrightarrow{t \ra \infty} 0} \ge 1 - \eta.
	\end{equation}	
\end{thm}

\begin{prf}

%	Let $\xi_t$ be the indicator of $q < \relprob$ at step $t$ and $\xi_0 := 1$. 
%	Notice that $\E{\sum_{i = 0}^{\infty}\xi_{i}} \leq \frac{1}{1 - \relfact}$, thus by Markov's inequality 
%	\begin{equation}
%		\forall C > 0 \spc \PP{\sum_{i = 0}^{\infty}\xi_{i} \geq C} \leq \frac{1}{C(1 - \relfact)},
%	\end{equation}		
%	which trivially implies
%	\begin{equation}
%		\PP{\sum_{i = 0}^{\infty}\xi_{i} = \infty} = 0.
%	\end{equation}
%    Now, let $\Omega^{\dagger} := \{ \omega \in \Omega \ | \ \sum_{i = 0}^{\infty}\xi_{i}[\omega] \neq \infty \}$.
%	From this point on throughout this proof it is assumed that we are working in the probability space induced on $\Omega^{\dagger}$.    
    
    % IF WE ALLOW RELAX ACTION AT STEP ZERO, THEN XI_0 MIGHT BE ZERO, BUT THE BOUND STILL HOLDS (IT'S CONSERVATIVE)
    
	Recalling \Cref{alg_calfq}, let us denote:
	\begin{equation}
	    \label{eqn_qdagger}
	    \begin{aligned}
	        & \hat Q^\dagger := \hat Q^{w^\dagger}(\state^\dagger).
	    \end{aligned}
	\end{equation}
	
	Next, we introduce:
	\begin{equation}
	    \label{calftimes}
	    \begin{aligned}
	        & \hat{\mathbb T}(\omega) := \{t \in \Z_{\ge 0} : \text{successful critic update} \}, \\
	        % & \tilde{\mathbb T}(\omega) := \{t \in \Z_{\ge 0} : \xi_{t}[\omega] = 1 \}, \\
	        % & \mathbb T^{\policy_0}(\omega) := \{0\} \cup \\
	        % & \qquad \left\{t \in \N : \begin{array}{l}
	        % (\text{successful critic update at } t-1 \lor \xi_{t - 1}[\omega] = 1) \spc \land \\
	        % \text{unsuccessful at } t 
	        % \end{array}
	        %\right\}.
	    \end{aligned}
	\end{equation}
	The former set represents the time steps at which the critic succeeds and the corresponding induced action will fire at the next step.
	The latter set consists of the time steps after each of which the critic first failed, discarding the subsequent failures if any occurred.
	
	%\green{Since mathbb T pi 0 was changed, the below definition also changed, now saying ``otherwise''}
	
	Now, let us define:
	\begin{equation*}
	    %    \label{eqn_calf2notation1}
	    \begin{aligned}
	        & \hat Q^\dagger_t :=  \begin{cases}
	                            \hat Q_t, t \in \hat{\mathbb T},\\
	                            \hat Q^\dagger_{t-1}, \text{ otherwise}.
	                        \end{cases}
	    \end{aligned}
	\end{equation*}
%	and, similarly, $\hat Q^\dagger_t$ for ease of reference.
	
	Next, observe that there are at most
	\begin{equation}
	    \label{eqn_CALF2_critic_reaching}
	        \hat T := \max \left\{ \frac{ \hat Q^\dagger_0 - \bar \nu}{\bar \nu}, 0 \right\}
	\end{equation}
	critic updates until the critic stops succeeding and hence only $\policy_0$ is invoked from that moment on. Hence $\hat{\mathbb T}(\omega)$ is a finite set.
	%Furthermore, for $\omega \in \Omega^{\dagger}$ evidently $\tilde{\mathbb T}(\omega)$ is too a finite set.
	%Since ${\mathbb T}^{\policy_0}(\omega) \subset \tilde{\mathbb T}(\omega) \cup \hat{\mathbb T}(\omega)$ it can be concluded that ${\mathbb T}^{\policy_0}(\omega)$ is a finite set.
	Notice $\hat T$ was independent of $\G'$ and in turn dependent on the initial value of the critic.

	%Consider some $t^{\dagger}$, a time step after which the critic failed to find a solution.
	%At step $t^\dagger+1$, the action $\action_{t^\dagger}$ is taken leading the state to transition into some $\state_{t^\dagger+1} = \transit(\state_{t^\dagger}, \action_{t^\dagger})$.
	%Now, either the critic finds a solution $w^\dagger_{t^\dagger+1}$ again, or $\policy_0$ is invoked.
	
	%In the latter case, by \eqref{eqn_goallim}, let $T^{\policy_0}(s_{t^\dagger+1}, \omega)$ be \sut
	%\[
	%	\forall t \ge t^\dagger+1 + T^{\policy_0}(s_{t^\dagger+1}, \omega) \spc \spc \goaldist(\State'_t(s_{t^\dagger+1},\omega)) \le h.
	%\]
	Let $t^\dagger_{\text{final}}[\omega] = \sup \hat{\mathbb T}[\omega]$. 
	Now notice that since $\Action_{t^{\dagger}_{\text{final}} + t + 1} \sim \policy_0(\State^\pi_{t^{\dagger}_{\text{final}} + t}(\state_{0}))$, by assumptions of the theorem it holds that 
	\begin{equation}
	\forall \tau \geq 0 \spc \PP{\goaldist(\State_{t^{\dagger}_{\text{final}} + t}^{\pi}) \xrightarrow{t \ra \infty} 0 \spc | \spc t^{\dagger}_{\text{final}} = \tau} \geq 1 - \eta.	
	\end{equation}
	Now, evidently by the law of total probability
	\begin{multline}
		\PP{\goaldist(\State_{t^{\dagger}_{\text{final}} + t}^{\pi}) \xrightarrow{t \ra \infty} 0} = \\ \sum_{\tau=0}^{\infty}\PP{\goaldist(\State_{t^{\dagger}_{\text{final}} + t}^{\pi}) \xrightarrow{t \ra \infty} 0 \spc | \spc t^{\dagger}_{\text{final}} = \tau}\PP{t^{\dagger}_{\text{final}} = \tau} \geq \\ (1 - \eta)\sum_{\tau=0}^{\infty}\PP{t^{\dagger}_{\text{final}} = \tau} = 1 - \eta.	
	\end{multline}
	
	This concludes the proof.
\end{prf}

From the statement of the theorem the following becomes clear:
\begin{center}
\textit{CALF reaches the goal even during \underline{online} learning.}
\end{center}

\section{Case study}
\label{sec_results}

We investigated the performance of CALF by benchmarking against other methods such as a Model Predictive Control (MPC), a modified version of State–action–reward–state–action (SARSA-m), a Proximal Policy Optimization (PPO), and the nominal policy $\policy^0$ provided to CALF as the baseline. 
These policies as well as the rest of the computational set up were implemented using Regelum \cite{regelum2024}, a Python framework for reinforcement learning, control and simulation. 
The best results of respective algorithms were reproduced on a wheeled mobile robot. 

SARSA-m is an algorithm obtained by removing line \ref{algline_recovery}: from \Cref{alg_calfq} \cite{Osinenko2024}.
The comparison with this algorithm is necessary to verify the significance of knowledge transfer between $\policy^0$ and the critic of CALF.
We observed no success with plain SARSA, hence the considered modification.

\emph{The results of experiments can be reproduced from code available via the following link:} \small \blue{\href{https://github.com/thd-research/calf-rl-mobile-robot/tree/migrate}{https://github.com/thd-research/calf-rl-mobile-robot/tree/migrate}}\normalsize.

\subsection{Environment description}

\begin{figure}[!h]
	\centering
	\includegraphics{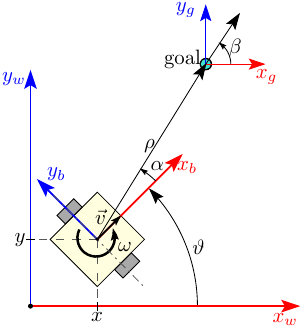}
	\caption{WMR kinematics and its frames of interests.}
	\label{fig_3wrobot}
\end{figure}

\begin{figure}[H]
	\centering
	\includegraphics[width=\linewidth]{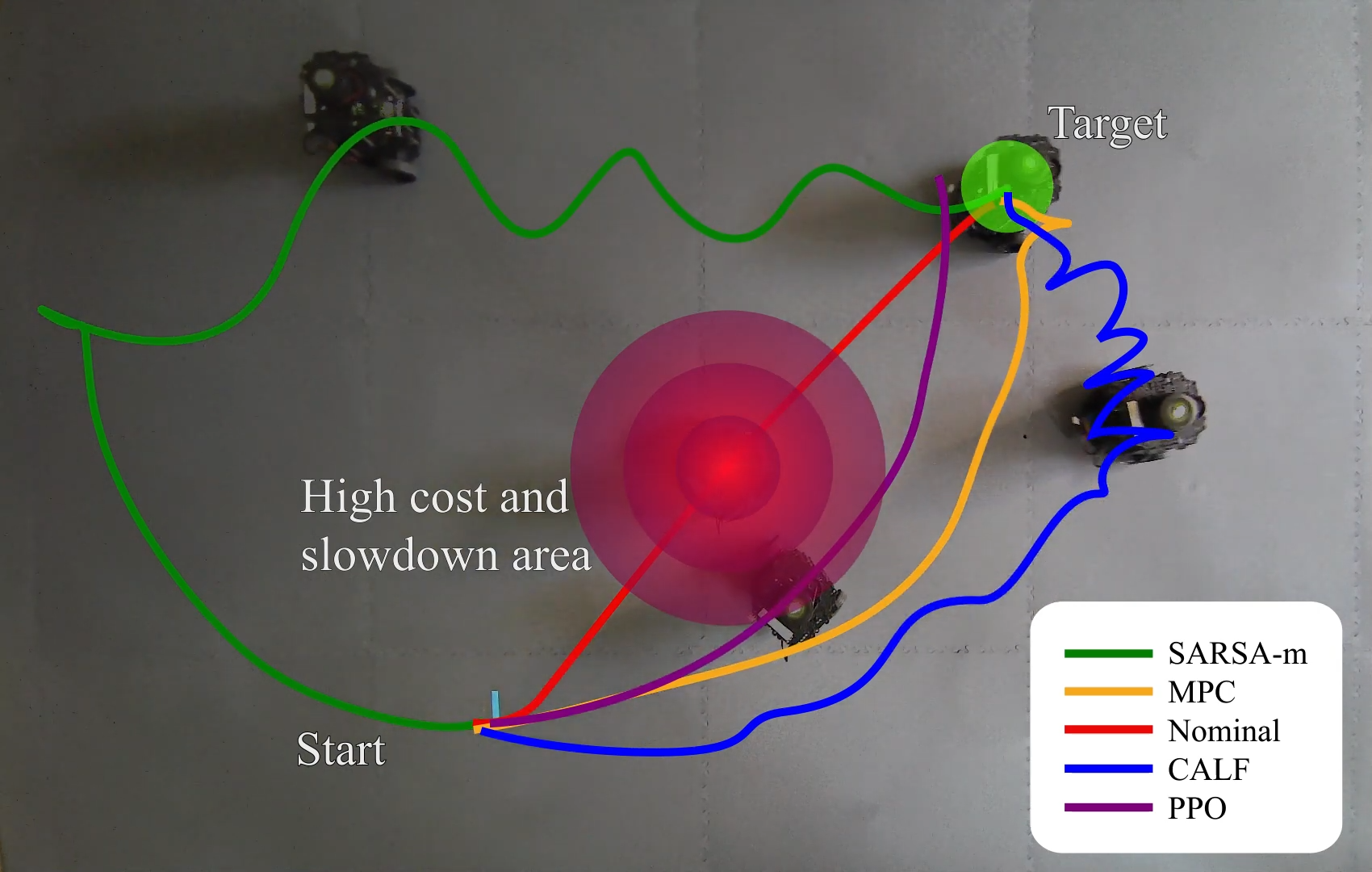}
	\caption{Real-world runs reproducing best episodes of respective algorithms (overlayed).
		The full footage can be found via the following link: \blue{\href{https://youtu.be/RgiDHzE5-w8?si=Kig6bNl8Cd7dTrzP}{https://youtu.be/RgiDHzE5-w8?si=Kig6bNl8Cd7dTrzP}}.}
	\label{fig_testbed}
\end{figure}

The proposed algorithms were tested by using the nonholonomic wheeled mobile robot (WMR) TurtleBot3 Burger by Robotis, schematically depicted in Fig. \ref{fig_3wrobot},
The corresponding testbed with selected result demonstration is shown in Fig.~\ref{fig_testbed}.
% The differential drive model characterizes the motion of a wheeled robot, with two actuated wheels, each capable of rotating independently of the other.
The idealistic differential equations that describe the dynamics of the robot (and do not account for uncertainty) read:
\begin{equation}
	\begin{aligned}
		\dot{x} &= v \cos(\vartheta), \\
		\dot{y} &= v \sin(\vartheta), \\
		\dot{\vartheta} &= \omega,
	\end{aligned}
\end{equation}
where $x$ and $y$ represent the coordinates of the robot in the plane, and \(\vartheta\) represents the yaw rotation of the robot relatively to facing the positive horizontal semi-axis, with the respective translational and rotational velocities $v, \omega$ (not to be confused with the sample variable $\omega$ from Section \ref{sec_theorems}).

%The goal of the control objective is to design a control strategy that allows the autonomous stabilization of the robot at the origin of a coordinate system, starting from non-zero initial conditions. 

\subsection{Objectives}

The environment has the following three points of interest: the initial state at $\State_{0} := (-1 \ -1 \ 0)^T$, the target $\mathbb G := \{(0 \ 0 \ 0)^T\}$ and the impeding area $\{(x, y, \vartheta)^{T} \ | \ (x + 0.5)^2 + (y + 0.5)^2 \leq 0.1^2\}$, where the speed of the robot is limited to 0.01 m/s due to \eg environmental conditions, irregular terrain or hazards.
%MPC10, MPC15 AND MPC25
%The robot's initial pose was set to [-1 -1 0] as [x y $\theta$], where $x$, $y$ are the coordinates of the robot position and $\theta$ the robot orientation recording to the world frame. The robot targets the goal [0 0 0]. The high-cost zone or ``impeding spot'' is located at ($x=-0.5$ and $y=-0.5$). This impeding spot restricts the robot by reducing its linear velocity (to 0.01 m/s) when it enters a radius of 0.1 meters around the center of this point; in addition, a penalty gain of 100 is assigned by a Gaussian distribution with the mean at the spot's location and the standard deviation of $0.1$.\\
%To evaluate these controllers equally and not prioritize axes, the coefficient diagonal matrix was set as [100, 100, 1, 0, 0], where each element stands for error in the x-axis, y-axis, theta, linear velocity, and angular velocity. Prioritizing any axis would introduce bias since the impeding spot is positioned directly along the path from the initial state to the goal pose.
The cost is designed in a way that penalizes both the displacement from the target and the proximity to the impeding area (the $\pi$ here is not to be confused with the policy):
\begin{equation}
	\label{eqn_3wrobotcost}
	\cost \left(s, a \right) = 100x^2 + 100y^2 + \vartheta^2 + 100  \cost' \left(x, y \right), 
\end{equation}
where 
\begin{equation}
	\label{eqn_hotspot}
%	\begin{aligned}
	\cost'(x, y) = \frac{1}{2 \pi 0.1^2 } \exp \Bigg(- \frac{1}{2} \left( \frac{(x + 0.5)^2}{0.1 ^ 2}  + \frac{(y + 0.5) ^2}{0.1 ^ 2} \right) \Bigg).
%	\end{aligned}
\end{equation}

\subsection{Nominal policy}

%Nominal controller for turtlebot. \\

%We now proceed to the description of the nominal policy.
The action components of the nominal policy $v, \omega$ were determined using the polar coordinate representation of the WMR as per \cite{Astolfi1995Exponentialsta}.
%\begin{equation}
%	\begin{bmatrix}
%		\dot{\rho} \\
%		\dot{\alpha} \\
%		\dot{\beta}
%	\end{bmatrix} = \begin{bmatrix}
%		\pm \cos {\alpha} & 0 \\
%		\mp \frac{\sin{\alpha}}{\rho} & 1 \\
%		\pm \frac{\sin{\alpha}}{\rho} & 0
%	\end{bmatrix}     \begin{bmatrix}
%		\pm v \\
%		\omega
%	\end{bmatrix},
%\end{equation}

%where the top sign (plus or minus) holds if $\alpha \in (- \frac \pi 2, \frac \pi 2]$ and the bottom sign holds if $\alpha \in \left(-\pi, -\pi/2 \right]\cup \left(\pi/2, \pi \right]$.
The transformation to polar coordinates reads:
\begin{align*}
	\rho &= \sqrt{x ^ 2 + y ^ 2},\\
	\alpha &= -\vartheta + \arctan2(y, x),\\
	\beta &= -\vartheta - \alpha.
	\label{eqn_polar}
\end{align*} 

The actions were set as per:
\begin{align}
	v \la \kappa_{\rho} \rho, \qquad 
	\omega \la \kappa_{\alpha}\alpha + \kappa_{\beta} \beta,
\end{align}
where $\kappa_{\rho} > 0, \kappa_{\beta} < 0, \kappa_{\alpha} - \kappa_{\rho} > 0$.

\subsection{Hyper-parameters and pre-training}

Excluding the nominal policy and MPC, the agents of CALF, SARSA-m, and PPO were trained over at least 40 episodes, with each episode composed of 50 time units of operation (at 0.1 s controller sampling period). 
All of the hyper-parameters used for studied approaches are illustrated in Table \ref{table_ctrl_param}.
%\begin{center}
%\textit{
For the sake of fair comparison, SARSA-m and PPO were \underline{pre-trained} in such a way that their performance at episode zero approximately matches that of the nominal policy.
However, such a pre-training does not guarantee goal reaching afterwards.
%}
%\end{center}

\section{Results and discussion} 

\begin{figure*}[t]
	\vspace{0.4cm}
     \begin{subfigure}[t]{0.48\textwidth}
         % \centering
         \includesvg[height=0.55\linewidth]{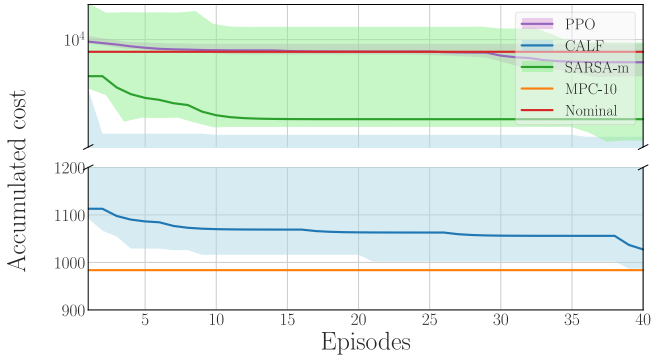}
         \caption{Total accumulated cost vs. episode.
         	Solid lines are median values of top 25\% best episodes, shaded: confidence intervals.}
%         	The costs attained by MPC and the nominal policy are provided for reference.}
     \end{subfigure}
     \hfill
     \begin{subfigure}[t]{0.48\textwidth}
         % \centering
         \includesvg[height=0.55\linewidth]{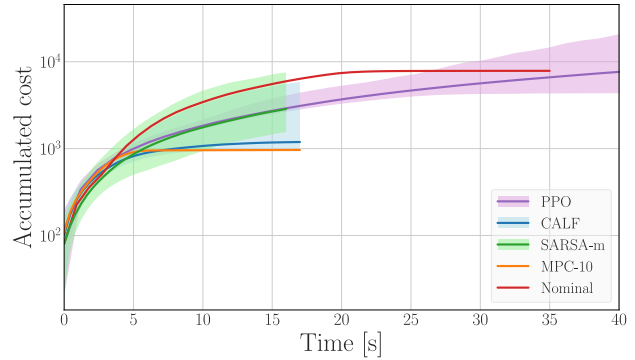}
         \caption{Accumulated cost vs. time (within episodes) with 95\% confidence intervals (solid: median).
%         	Solid lines are median values.
         	The plots are shown until goal reaching was achieved.}
     \end{subfigure}
        \caption{Learning curves obtained from 20 seeds [1..20].}
        \label{fig_cost}
\end{figure*}

\begin{figure*}[t]
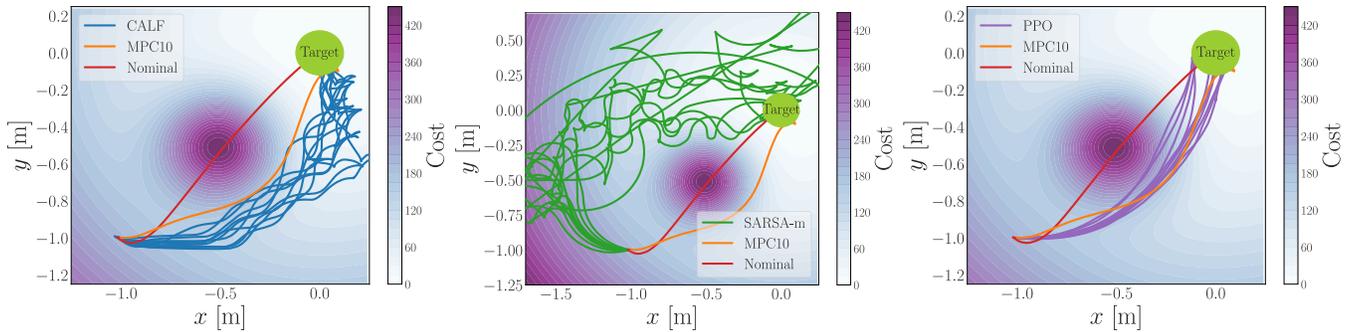

     \begin{subfigure}[b]{0.33\textwidth}
         \centering
         \includesvg[width=\linewidth]{gfx/report_CALF_trajectory.svg}
%         \caption{CALF and references}
     \end{subfigure}
     % \hfill
     \begin{subfigure}[b]{0.33\textwidth}
         \centering
         \includesvg[width=\linewidth]{gfx/report_SARSA-m_trajectory.svg}
%         \caption{SARSA-m and references.}
     \end{subfigure}
     % \hfill
     \begin{subfigure}[b]{0.33\textwidth}
         \centering
         \includesvg[width=\linewidth]{gfx/report_PPO_trajectory.svg}
%         \caption{PPO and references.}
     \end{subfigure}
        \caption{Trajectories in best episodes over 20 random seeds.
        	The succeess rate of goal reaching was 100 \% of episodes by CALF, as expected, not more than 30 \% by PPO, and about 50 \% by SARSA-m.
%        	 It should be noted that CALF reached the goal in every episode, not just in the depicted episodes, unlike PPO and SARSA-m.
%        Overall, the success rate of goal reaching in our experiments did not exceed 30 \% of episodes by PPO, and higher by SARSA-m.}
}
        \label{fig_traj}
\end{figure*}

\begin{table*}[t]
	% \centering
	% \noindent\begin{minipage}{\textwidth}
		\centering
		\begin{tabular}{ |c|c|c|c|c|c| } 
			\hline
			& CALF & MPC & SARSA-m & PPO & Nominal \\
			\hline
			%\makecell{sampling time \\ dt [s]} & 0.1 & 0.1 & 0.1 & 0.1 & 0.1\\ 
			% \hline
			% \makecell{objective \\ coefficient} & 100, 100, 1, 0, 0 & 100, 100, 1, 0, 0 & 100, 100, 1, 0, 0 & 100, 100, 1, 0, 0 & 100, 100, 1, 0, 0\\ 
			%\hline
			\multirow{3}{4em} {\makecell{Policy \\ parameters}} 
			& $\pi_{0}$: Nominal & horizon length:  & $\pi_{0}$: Null &  hidden layer: 2   & $\kappa_{\rho}$: 0.2     \\ 
			&                    & 10       &                 &  dimension: 15     & $\kappa_{\alpha}$: 1.5   \\ 
			&                    & step size: 4  &                 & lr: 5e-3              & $\kappa_{\beta}$: -0.15 \\ 
			%    &                    &                  &                 & discount factor: 0.9 &               \\ 
			&                    &                  &                 & clip range: 0.2      &               \\ 
			%    &                    &                  &                 & lr:  5e-3  \red{?}          &               \\ 
			%    &                    &                  &                 & GAE: No use          &               \\ 
			%    &                    &                  &                 &  epoch/step: 50    &               \\ 
			
			Discount rate    & $\gamma: 0.9$ & $\gamma: 1.0$ & $\gamma: 0.9$ & $\gamma: 0.9$ &   \\
			\hline
			\multirow{2}{4em}{Critic} & $\bar{\nu}: 1e-6$      &  & $\bar{\nu}: 1e-6$           &  hidden layers: 3 &  \\ 
			& $\bar{\nu}_{\text{max}}: 0.1$ &  & $\bar{\nu}_{\text{max}}: 0.1$ &  layer dimension: 15   &  \\ 
			& $\hat{\kappa}_{\text{low}}: 0.1$   &  & $\hat{\kappa}_{\text{low}}: 0.1$   & lr: 0.1            &  \\ 
			& $\hat{\kappa}_{\text{up}}: 1e3$    &  & $\hat{\kappa}_{\text{up}}: 5e2$    &  steps/episode: 50  &  \\
			
			Critic structure & $w_{1}x + w_{2}y + w_{3}\vartheta$    &       -       &  $w_{1}x + w_{2}y + w_{3}\vartheta$   &       multilayer perceptron       &       -       \\
			\hline
			%\makecell{\# learning \\ episodes} & 40 & - & 40 & 100 & - \\
			%\hline
			
		\end{tabular}
		
		\caption{Hyper-parameters of the controllers studied.}
		\label{table_ctrl_param}
	\end{table*}

Fig. \ref{fig_cost} and Fig. \ref{fig_traj} depict the learning curves and best performing trajectories respectively. 
It was not surprising that MPC with such a fairly large horizon (10) produced near optimal results, which is precisely the motivation behind benchmarking against it. 
CALF reached the goal quicker and maintained a safe distance from the impeding spot. 
It should be remarked again that CALF is model-free whereas MPC with long horizons may be computationally expensive.
PPO in turn showed the potential to learn an optimal path to the goal while avoiding the impeding spot. 
Still, PPO did not fully utilize the robot's maximum velocity within the limited number of learning episodes and sometimes did not successfully reach the target. 
On the other hand, SARSA-m generally succeeded in maintaining a safe distance from the impeding spot but often displayed an erratic behavior when approaching the goal.
%The nominal controller showed the poorest performance when crossing the impeding spot.
Three important observations can be made: \textbullet~both PPO and SARSA-m outperform the nominal baseline upon learning convergence, \textbullet~CALF outperforms both SARSA-m and PPO, \textbullet~the performance of CALF is close to MPC with large horizon (MPC).
%\begin{enumerate}
%\item Both PPO and SARSA-m outperform the nominal baseline upon learning convergence,
%\item CALF outperforms both SARSA-m and PPO,
%\item The performance of CALF is close to MPC with large horizon (MPC).
%\end{enumerate}

Since CALF, PPO and SARSA-m had equal head starts it would be fair to conclude that CALF is superior in terms of sample efficiency in the considered few-episode setting. 
Notice that even in the first few episodes CALF tends to immediately make use of the information extracted from the nominal policy and get far ahead both the nominal baseline itself and the other RL algorithms considered in the study. 
Unlike PPO, CALF did not missed the target even once, despite somewhat wider confidence bounds.

\section{Conclusion}
%\textbf{Concluding remarks}.
This work proposes a novel RL algorithm with guarantees for stochastic environments that combines learning with a system of Lyapunov-like constraints. 
The resulting approach named Critic As a Lyapunov Function (CALF) possesses goal reaching guarantees that were rigorously formulated and proven. 
An empirical study with a real non-holonomic WMR was conducted together with a set of experiments that compare the approach to state-of-the-art RL algorithms. 
Overall the outcomes of the experiments indicate that CALF is superior in terms of sample efficiency in the considered few-episode setting.

%%%%%%%%%%%%%%%%%%%%%%%%%%%%%%%%%%%%%%%%%%%%%%%%%%%%%%%%%%%%%%%%%%%%%%%%%%%%%%%%%%%%%%%%%%%%
%%%%%%%%%%%%%%%%%%%%%%%%%%%%%%%%%%%%%%%%%%%%%%%%%%%%%%% BIBLIOGRAPHY

\bibliographystyle{IEEEtran}
\bibliography{
bib/Osinenko__Dec2023,
bib/AIDA__Mar2024,
bib/Ruben_SOTA,
bib/Turtlebot__Sep2024
}

%\clearpage
%\raggedbottom
%\pagebreak

\end{document}